\documentclass{article}

\usepackage[preprint]{neurips_2026}

\usepackage[utf8]{inputenc} 
\usepackage[T1]{fontenc}    
\usepackage{hyperref}       
\usepackage{url}            
\usepackage{booktabs}       
\usepackage{amsfonts}       
\usepackage{nicefrac}       
\usepackage{microtype}      
\usepackage{xcolor}         

\usepackage{amsmath}
\usepackage{amssymb}
\usepackage{mathtools}
\usepackage{amsthm}
\usepackage{adjustbox}
\usepackage{wrapfig}
\usepackage{pifont}
\usepackage{subcaption}
\usepackage{graphicx}

\usepackage[capitalize,noabbrev]{cleveref}

\theoremstyle{plain}

\theoremstyle{definition}

\theoremstyle{remark}

\usepackage{tikz}
\usepackage{pgfplots}
\pgfplotsset{compat=1.18}

\usepackage{relsize}

\usepackage{multirow}

\title{MemDLM: Memory-Enhanced DLM Training}

\author{Zehua Pei$^1$,  
Hui-Ling Zhen$^2$, 
Weizhe Lin$^2$, 
Sinno Jialin Pan$^1$, \\
\bf Yunhe Wang$^2$,
Mingxuan Yuan$^2$, 
Bei Yu$^1$\\
$^1$The Chinese University of Hong Kong \quad
$^2$Huawei Technologies Co., Ltd
}

\begin{document}

\maketitle

\begin{abstract}
Diffusion Language Models (DLMs) offer attractive advantages over Auto-Regressive (AR) models, such as full-attention parallel decoding and flexible generation. 
However, standard DLM training uses a static, single-step masked prediction objective that never exposes the model to the progressive denoising dynamics of inference, and forces all contextual information to be maintained purely through token-space attention, which becomes increasingly diluted as context length grows.
We propose \textbf{MemDLM} (Memory-Enhanced DLM), which introduces a second memory channel by embedding a simulated denoising trajectory into training via Bi-level Optimization. 
An inner loop updates a set of fast weights, forming a \emph{Parametric Memory} that captures the local trajectory experience, while an outer loop updates the base model conditioned on this memory. 
By offloading part of the memorization burden from token-space attention to parameter space, MemDLM yields faster convergence, stronger long-context representations, and lower training loss, even when the fast weights are discarded at inference time. 
Re-enabling the inner loop at inference provides an additional prompt-specific adaptation effect, where the Parametric Memory acts as an emergent \emph{in-weight retrieval} mechanism on challenging Needle-in-a-Haystack tasks. 
\end{abstract}

\begin{figure}[h]
    \centering
    \includegraphics[width=0.95\textwidth]{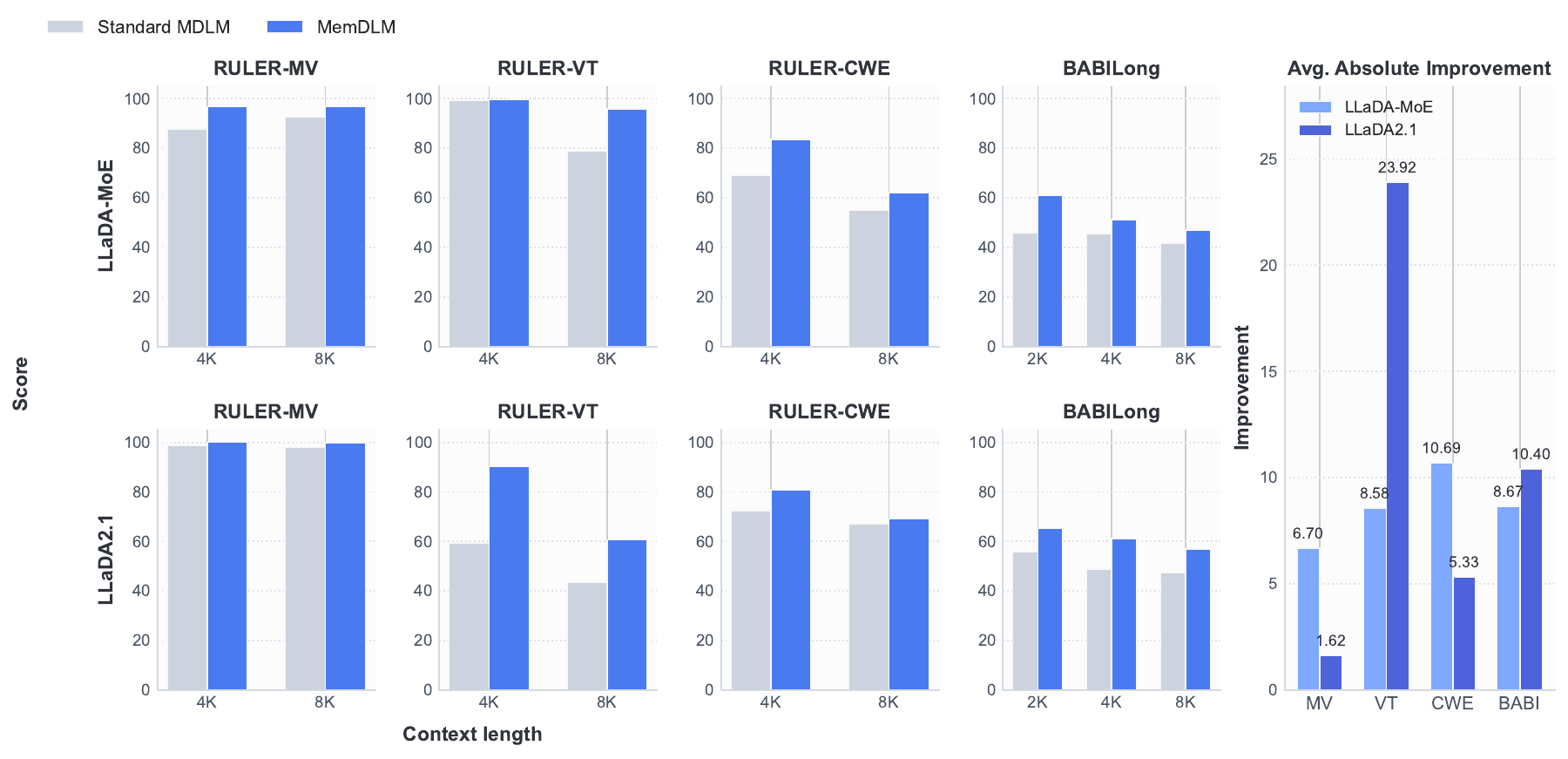}
    \caption{\textbf{Needle-in-a-Haystack results overview.} Gray bars denote Standard MDLM and blue bars denote MemDLM. Left: detailed results on RULER-MV, RULER-VT, RULER-CWE, and BABILong for the LLaDA-MoE-7B-A1B-Base and LLaDA2.1-mini backbones. Right: mean absolute improvement of MemDLM over Standard MDLM for each task, averaged across the evaluated context lengths within each backbone.}
    \label{fig:teaser}
\end{figure}
\section{Introduction}

Diffusion Language Models (DLMs) are a promising alternative to Auto-Regressive (AR) models, offering parallel generation, bidirectional context awareness, and flexible text manipulation~\cite{austin2021structured, sahoo2024simple, lou2023discrete, shi2024simplified, ou2024your, zheng2024masked, campbell2022continuous, sun2022score, meng2022concrete}. 
However, standard DLM training uses a static Masked Diffusion Language Modeling (MDLM) objective that never exposes the model to the progressive denoising dynamics encountered at inference~\cite{he2025mdpo, wang2025revolutionizing, huang2025reinforcing, peng2025planner}, and forces all contextual information to be maintained purely through token-space attention, which becomes increasingly diluted as context length grows~\cite{nakanishi2025scalable, ye2026dysco, liu2024lost, xiao2023efficient}.

We propose \textbf{MemDLM} (Memory-Enhanced DLM), which introduces a second memory channel by writing local denoising trajectory information into parameter space during training. 
An inner optimization loop simulates progressive denoising and updates parameter-efficient fast weights, which act as a \emph{Parametric Memory} of the local trajectory~\cite{tieleman2009using, ba2016using, hinton1987using, sprechmann2018memory}. 
Because part of the memorization burden is offloaded from attention to these fast weights, the base model is freed from preserving everything through token-space representations alone and is simultaneously exposed to richer trajectory structure than the static single-step objective provides.

\begin{figure*}[t]
    \centering
    \vspace{-0.5cm}
    \includegraphics[width=\textwidth]{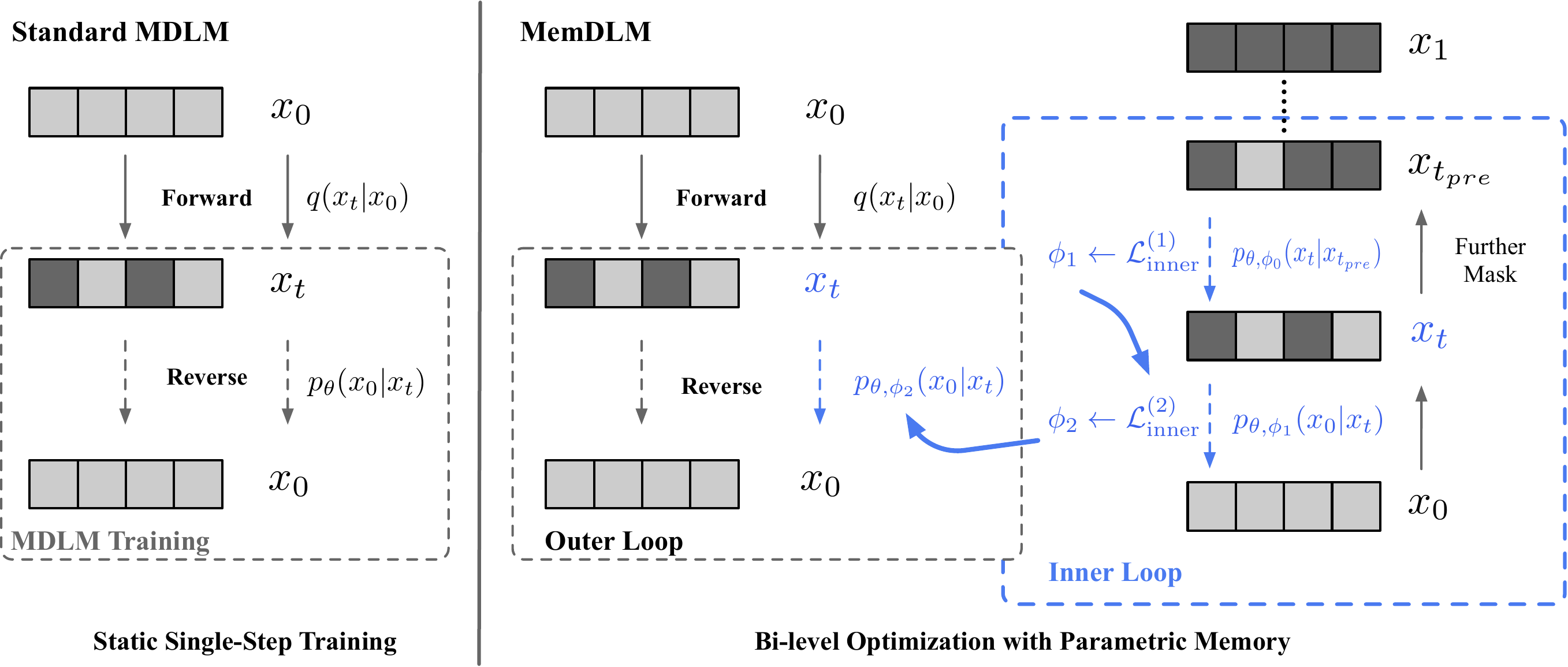}
    \caption{Overview of MemDLM. 
    \textbf{Left:} standard MDLM training uses a static single-step denoising objective from $x_t$ to $x_0$. 
    \textbf{Right:} MemDLM uses Bi-level Optimization in which an inner loop updates fast weights $\phi$ along an anchor-consistent local trajectory ($x_{t_{\textit{pre}}} \rightarrow x_t \rightarrow x_0$), and the outer loop updates the base model $\theta$ on the anchor state $x_t$ conditioned on this parametric memory. 
    \textbf{Legend:} dark tokens denote mask tokens, light tokens denote observed tokens, straight arrows denote forward or reverse prediction flow, and blue curved arrows denote inner-loop fast-weight updates.}
    \label{fig:overview}
    \vspace{-0.5cm}
\end{figure*}

\Cref{fig:overview} summarizes this design. By internalizing local trajectory information into transient fast weights, MemDLM reduces the memorization pressure on attention and enriches the training signal, yielding stronger long-context representations even when the fast weights are discarded at inference time. When the inner loop is optionally re-enabled at inference, it provides an additional prompt-specific adaptation pathway.
Empirically, on LLaDA-MoE~\cite{zhu2025llada}, MemDLM improves RULER Variable Tracking~\cite{hsieh2024ruler} at 8K from $78.8\%$ to $95.8\%$, and on LLaDA2.1~\cite{bie2026llada2}, it improves BABILong~\cite{kuratov2024babilong} at 8K from $47.4\%$ to $57.0\%$.

In summary, our contributions are:
\begin{itemize}
    \item We identify two limitations of standard DLM training: the static objective never exposes the model to progressive denoising dynamics, and token-space attention alone becomes insufficient for preserving task-relevant information in long contexts.
    \item We introduce \textbf{MemDLM}, a Bi-level Optimization framework that simulates progressive denoising during training and induces a Parametric Memory mechanism, creating a second memory channel that complements attention.
    \item We show that this memory-aware training improves optimization and long-context performance even when the fast weights are discarded at inference time, while re-enabling the inner loop at inference provides an additional prompt-specific adaptation effect that we interpret as \emph{in-weight retrieval}.
\end{itemize}

\section{Preliminaries and Motivation}
\label{sec:motivation}

Before formalizing our method, we review the standard training and inference paradigms of Masked Diffusion Language Models (MDLMs)~\cite{sahoo2024simple, shi2024simplified} and identify two limitations of this paradigm.

\subsection{Preliminaries: Masked Diffusion Language Models}
Consider a sequence of clean text comprising $L$ tokens, denoted as $x_0 = (x_0^1, \dots, x_0^L)$, where each token belongs to a discrete vocabulary $\mathcal{V}$. Discrete diffusion models operate by defining a forward corruption process that gradually introduces noise over a continuous time variable $t \in [0, 1]$. At $t=0$, the sequence is completely clean ($x_0$), and at $t=1$, the sequence reaches a state of pure noise ($x_1$). The model is then trained to approximate the reverse generative process, learning to map a noisy state $x_t$ back to the original text $x_0$.

\textbf{Absorbing-State Masking.} In the specific framework of MDLMs, the forward corruption $q(x_t | x_0)$ is instantiated as an absorbing-state process. Rather than transitioning tokens to random vocabulary items, tokens are replaced by a dedicated absorbing token, $m \notin \mathcal{V}$ (often denoted as \texttt{[MASK]}). Under a linear noise schedule, the probability that the $i$-th token is masked at time $t$ is simply $t$:
\begin{equation}
    q(x_t^i | x_0^i) = (1 - t) \mathbb{I}(x_t^i = x_0^i) + t \mathbb{I}(x_t^i = m),
\end{equation}
where $\mathbb{I}(\cdot)$ denotes the indicator function. 

\textbf{Training via Static Masking.} The objective of the neural network $p_\theta(x_0 | x_t)$, parameterized by $\theta$, is to reconstruct the clean tokens $x_0$ given the corrupted sequence $x_t$. Because unmasked tokens are perfectly preserved in the absorbing-state formulation, the model only needs to predict the identities of the tokens at the currently masked indices, $\mathcal{M}_t = \{i \mid x_t^i = m\}$. 

Standard MDLM training minimizes the expected negative log-likelihood of these masked tokens over uniformly sampled timesteps, yielding the following objective:
\begin{equation}
    \mathcal{L}_{\text{MDLM}}(\theta) = \mathbb{E}_{t \sim \mathcal{U}(0,1), x_0} \left[ \omega(t) \sum_{i \in \mathcal{M}_t} -\log p_\theta(x_0^i | x_t) \right],
    \label{eq:standard_mdlm}
\end{equation}
where $\omega(t)$ serves as a time-dependent weighting factor (e.g., $\omega(t) = 1/t$) to balance the loss across varying noise levels. Critically, \cref{eq:standard_mdlm} represents a single-step, static masking objective: the model receives a masked text based purely on ground-truth data and is optimized to predict the clean sequence in one isolated step.

\textbf{Inference via Iterative Denoising.} In contrast, DLMs generate text during inference through a multi-step, progressive denoising trajectory. Starting from a fully masked sequence at $t=1.0$, the model predicts the clean tokens. A subset of the highest-confidence predictions is then unmasked to form a partially noisy intermediate sequence $x_{t - \Delta t}$. This process repeats iteratively until $t=0$, where all tokens are decoded. Crucially, at each step, the model's input is conditioned on its own noisy predictions from previous steps, rather than pristine ground-truth context.

\subsection{Motivation}

Standard MDLM training has two limitations.

\textbf{Train-inference mismatch.} The static single-step objective in \cref{eq:standard_mdlm} never exposes the model to the progressive denoising dynamics it encounters at inference time. This discrepancy, widely documented as exposure bias in DLMs~\cite{he2025mdpo, wang2025revolutionizing, huang2025reinforcing, peng2025planner}, means that representations learned during training are not aligned with the sequential, multi-step process the model actually executes. We provide an empirical quantification of this mismatch in \cref{app:exposure_bias}.

\textbf{Attention dilution in long contexts.} A second, complementary challenge arises as context length grows. Dense softmax attention must distribute probability mass across all tokens in the sequence. 
As the number of tokens increases, each individual token receives a smaller share of attention, making it harder for the model to focus on task-relevant information~\cite{nakanishi2025scalable, ye2026dysco, liu2024lost, xiao2023efficient}. 
Standard MDLM training offers no mechanism to mitigate this, as it forces the model to maintain all task-relevant contextual information purely through these increasingly diluted token-space attention patterns.

These observations motivate introducing a second, complementary memory channel alongside token-space attention. If part of the local denoising trajectory can be written into parameter space during training, the base model is exposed to richer trajectory structure than the static single-step objective provides, while the memorization burden on the attention mechanism is reduced. As we show empirically, both effects contribute to stronger long-context representations, even when the parametric memory is discarded at inference time.

\section{Methodology}
\label{sec:method}

Motivated by the limitations identified in \cref{sec:motivation}, we aim to introduce a second memory channel that complements token-space attention. We achieve this by proposing \textbf{MemDLM}, which embeds a simulated denoising trajectory into training via a Bi-level Optimization framework that naturally induces a Parametric Memory mechanism.

\subsection{Bi-level Optimization for Denoising Simulation}
To align the training objective with the iterative nature of inference, we partition the model parameters into the base weights $\theta$ and a set of parameter-efficient fast weights $\phi$ (e.g., low-rank adapters). We formulate the training process as a Bi-level Optimization problem:
\begin{align}
    \min_\theta \quad & \mathbb{E}_{t \sim \mathcal{U}(0,1), x_0} \left[ \omega(t) \sum_{i \in \mathcal{M}_t} -\log p_{\theta, \phi_K}(x_0^i \mid x_t) \right], \label{eq:outer_objective} \\
    \text{subject to} \quad & \phi_k = \phi_{k-1} - \eta \nabla_\phi \mathcal{L}_{\text{inner}}^{(k)}(\theta, \phi_{k-1}) \quad \text{for } k=1, \dots, K. \label{eq:inner_objective}
\end{align}

Here, \cref{eq:inner_objective} represents the inner loop, which simulates an unrolled $K$-step denoising trajectory for a specific batch. Starting from initial zero weights $\phi_0 = \mathbf{0}$, the fast weights dynamically accumulate sample-specific contextual details through gradient descent, resulting in a final state $\phi_K$ that acts as a \emph{Parametric Memory} of the local trajectory experience. \Cref{eq:outer_objective} represents the outer loop, where the base model $\theta$ is updated conditioned on this internalized memory.

\subsection{The Inner Loop: Anchor-Consistent Trajectories}
Rather than applying an arbitrary sequence of masks, we design the inner loop to simulate an \textbf{Anchor-Consistent Local Trajectory}. Because the outer objective is computed exactly at the noisy state $x_t$, the inner loop's parametric memory is most effective when it explicitly targets and processes this exact anchor state.
This kind of masked inner-loop refinement is especially natural for DLMs: bidirectional denoising lets the model aggregate information from all visible tokens while updating multiple masked positions in a single step, whereas comparable hole-filling supervision is less direct under standard left-to-right auto-regressive factorization.

We formulate the inner loop as a two-stage gradient update ($K=2$), initializing the fast weights to zero ($\phi_0 = \mathbf{0}$). In the first stage (\textbf{Pre-Anchor Alignment}), we construct a noisier local state $x_{t_{\text{pre}}}$ (where $t_{\text{pre}} > t$) by further masking the anchor state $x_t$. The model then denoises $x_{t_{\text{pre}}}$ toward the anchor state $x_t$. In the second stage (\textbf{Anchor-to-Target}), the model takes the exact anchor state $x_t$ and predicts the final clean state $x_0$. 

The fast weights accumulate the trajectory dynamics through the following sequence of updates:
\begin{align}
    \mathcal{L}_{\text{inner}}^{(1)} &= \sum_{i \in \mathcal{M}_{t_{\text{pre}}}} -\log p_{\theta, \phi_0}(x_t^i \mid x_{t_{\text{pre}}}), \label{eq:inner_loss1} \qquad
    \phi_1 = \phi_0 - \eta \nabla_\phi \mathcal{L}_{\text{inner}}^{(1)}, \\
    \mathcal{L}_{\text{inner}}^{(2)} &= \sum_{i \in \mathcal{M}_t} -\log p_{\theta, \phi_1}(x_0^i \mid x_t), \label{eq:inner_loss2} \qquad
    \phi_2 = \phi_1 - \eta \nabla_\phi \mathcal{L}_{\text{inner}}^{(2)},
\end{align}
where $\eta$ is the inner learning rate. Together, these two stages encourage the fast weights to capture how a noisier local state transitions through the anchor state $x_t$ toward the clean target $x_0$. In this way, the inner loop accumulates an anchor-centered local trajectory in the final parametric state $\phi_2$.

\subsection{The Outer Loop: Conditioned Denoising}
After the inner loop accumulates the adapted parameters $\phi_2$ for a given batch, the outer objective is computed on the exact same anchor timestep $t$ and masked state $x_t$. The full outer objective mirrors standard MDLM training, but conditions the prediction on the Parametric Memory $\phi_2$:
\begin{equation}
    \mathcal{L}_{\text{MemDLM}}(\theta) = \mathbb{E}_{t \sim \mathcal{U}(0,1), x_0} \left[ \omega(t) \sum_{i \in \mathcal{M}_t} -\log p_{\theta, \phi_2}(x_0^i \mid x_t) \right].
\end{equation}

To update the base parameters $\theta$, we employ a First-Order approximation. This avoids the computationally prohibitive calculation of second-order Hessian matrices by treating the inner gradients $\nabla_\phi \mathcal{L}_{\text{inner}}$ as independent of $\theta$ during the outer backward pass. For a given training batch, the update rule for the base model is computed using the per-sample loss:
\begin{equation}
    \theta \leftarrow \theta - \beta \nabla_\theta \left( \omega(t) \sum_{i \in \mathcal{M}_t} -\log p_{\theta, \phi_2}(x_0^i \mid x_t) \right),
    \label{eq:outer_update}
\end{equation}
where $\beta$ is the outer learning rate. 
Because the fast weights $\phi_2$ can absorb part of the batch-specific trajectory information, the gradients $\nabla_\theta$ generated by \cref{eq:outer_update} may place less pressure on the base model to memorize local context purely in token space. This interpretation is consistent with the faster convergence and stronger downstream performance observed in our experiments.
\section{Experiments}
\label{sec:exp}


\subsection{Experimental Setup}
\label{sec:exp_setup}

We implement MemDLM in PyTorch~\cite{paszke2019pytorch} on top of \texttt{dllm}~\cite{zhou2026dllm} and evaluate with \texttt{lm-evaluation-harness}~\cite{eval-harness}. Main experiments use \texttt{LLaDA-MoE-7B-A1B-Base}~\cite{zhu2025llada} and \texttt{LLaDA2.1-mini}~\cite{bie2026llada2}, abbreviated as \textbf{LLaDA-MoE} and \textbf{LLaDA2.1}; unless noted otherwise, targeted analyses and ablations use LLaDA-MoE. We instruction-tune on LongAlpaca~\cite{long-alpaca} with asymmetric masking, use LoRA~\cite{hu2021lora} for both outer- and inner-loop adaptation, and evaluate on RULER~\cite{hsieh2024ruler}, BABILong~\cite{kuratov2024babilong}, and LongBench~\cite{bai2024longbench}. Full setup details are deferred to \cref{app:exp_setup}, and benchmark descriptions to \cref{app:benchmark_details}.

\subsection{Main Results: Long-Context Information Retrieval}
\label{sec:exp_retrieval}

Needle-in-a-Haystack retrieval is challenging for DLMs because the model must preserve a small amount of task-relevant information across long token-space contexts. We evaluate RULER (Multi-Value, Variable Tracking, and Common Words Extraction) and BABILong.

\begin{table*}[t]
    \centering
    \caption{Needle-in-a-Haystack results on RULER and BABILong. We report Standard MDLM, MemDLM (Train-Only), and MemDLM (Train \& Inference). Bold indicates the best result within each backbone block.}
    \label{tab:niah}
    \resizebox{\textwidth}{!}{
    \begin{tabular}{l l ccccccccc}
        \toprule
        \multirow{2}{*}{\textbf{Backbone}} & \multirow{2}{*}{\textbf{Method}} & \multicolumn{2}{c}{\textbf{RULER-MV}} & \multicolumn{2}{c}{\textbf{RULER-VT}} & \multicolumn{2}{c}{\textbf{RULER-CWE}} & \multicolumn{3}{c}{\textbf{BABILong}} \\
        \cmidrule(lr){3-4} \cmidrule(lr){5-6} \cmidrule(lr){7-8} \cmidrule(lr){9-11}
        & & \textbf{4K} & \textbf{8K} & \textbf{4K} & \textbf{8K} & \textbf{4K} & \textbf{8K} & \textbf{2K} & \textbf{4K} & \textbf{8K} \\
        \midrule
        \multirow{3}{*}{LLaDA-MoE} & Standard MDLM & $87.90$ & $92.50$ & $99.40$ & $78.84$ & $69.10$ & $55.18$ & $46.00$ & $45.60$ & $41.80$ \\
         & MemDLM (Train-Only) & $95.85$ & $96.95$ & $99.52$ & $94.84$ & $75.72$ & $58.20$ & $59.40$ & $50.60$ & $45.60$ \\
         & MemDLM (Train \& Inference) & $\mathbf{96.80}$ & $\mathbf{97.00}$ & $\mathbf{99.60}$ & $\mathbf{95.80}$ & $\mathbf{83.70}$ & $\mathbf{61.96}$ & $\mathbf{61.20}$ & $\mathbf{51.20}$ & $\mathbf{47.00}$ \\
        \midrule
        \multirow{3}{*}{LLaDA2.1} & Standard MDLM & $98.65$ & $98.00$ & $59.32$ & $43.72$ & $72.34$ & $67.08$ & $56.00$ & $48.80$ & $47.40$ \\
         & MemDLM (Train-Only) & $99.65$ & $99.00$ & $88.56$ & $59.15$ & $78.96$ & $68.02$ & $63.50$ & $58.50$ & $55.00$ \\
         & MemDLM (Train \& Inference) & $\mathbf{100.00}$ & $\mathbf{99.90}$ & $\mathbf{90.16}$ & $\mathbf{60.72}$ & $\mathbf{80.96}$ & $\mathbf{69.12}$ & $\mathbf{65.20}$ & $\mathbf{61.20}$ & $\mathbf{57.00}$ \\
        \bottomrule
    \end{tabular}
    }
\end{table*}

As shown in \cref{tab:niah}, MemDLM consistently improves over Standard MDLM across both backbones, especially at longer contexts. Crucially, the \textit{Train-Only} variant already yields large gains, indicating that the main benefit is induced during training rather than coming solely from re-running the inner loop at inference. For example, on LLaDA-MoE, MemDLM improves RULER Variable Tracking at 8K from $78.8\%$ to $95.8\%$, while on LLaDA2.1 it improves BABILong at 8K from $47.4\%$ to $57.0\%$. Re-enabling the inner loop at inference then provides an additional prompt-specific adaptation effect, which we interpret as \emph{in-weight retrieval}.

\paragraph{Length extrapolation via Parametric Memory.}
To probe robustness beyond the native 8K setting, we evaluate LLaDA-MoE at 16K and 32K. As shown in \cref{tab:niah_extrapolation}, performance drops for all methods as context grows, but MemDLM continues to improve over the baseline even in this extrapolation regime, suggesting that Parametric Memory helps preserve useful long-context representations beyond the lengths emphasized during training.

\begin{table*}[t]
    \centering
    \caption{Length extrapolation on Needle-in-a-Haystack tasks with the LLaDA-MoE backbone. MemDLM continues to outperform Standard MDLM at 16K and 32K across RULER and BABILong.}
    \label{tab:niah_extrapolation}
    \resizebox{0.83\textwidth}{!}{
    \begin{tabular}{l cccccccc}
        \toprule
        \multirow{2}{*}{\textbf{Method}} & \multicolumn{2}{c}{\textbf{RULER-MV}} & \multicolumn{2}{c}{\textbf{RULER-VT}} & \multicolumn{2}{c}{\textbf{RULER-CWE}} & \multicolumn{2}{c}{\textbf{BABILong}} \\
        \cmidrule(lr){2-3} \cmidrule(lr){4-5} \cmidrule(lr){6-7} \cmidrule(lr){8-9}
        & \textbf{16K} & \textbf{32K} & \textbf{16K} & \textbf{32K} & \textbf{16K} & \textbf{32K} & \textbf{16K} & \textbf{32K} \\
        \midrule
        Standard MDLM & $22.35$ & $14.30$ & $52.56$ & $9.48$ & $44.20$ & $17.82$ & $19.20$ & $6.80$ \\
        MemDLM (Train-Only) & $25.48$ & $15.20$ & $55.30$ & $10.35$ & $54.25$ & $22.48$ & $21.00$ & $8.50$ \\
        MemDLM (Train \& Inference) & $\mathbf{29.40}$ & $\mathbf{15.35}$ & $\mathbf{56.84}$ & $\mathbf{11.44}$ & $\mathbf{57.24}$ & $\mathbf{24.48}$ & $\mathbf{22.20}$ & $\mathbf{9.00}$ \\
        \bottomrule
    \end{tabular}
    }
\end{table*}

\subsection{Long-Context Generalization}
\label{sec:exp_longbench}

Building on the retrieval results, we evaluate MemDLM on LongBench under two settings: \textit{Train-Only} and \textit{Train \& Inference}.
As shown in \cref{tab:longbench}, Parametric Memory improves long-context task performance even in zero-shot \textit{Train-Only} mode, mirroring the NIAH results. Re-enabling the inner loop at inference yields consistent further gains, indicating that prompt-specific adaptation complements the training-time benefit.

\begin{table}[t]
    \centering
    \caption{LongBench results on LLaDA-MoE. \textbf{Train-Only} uses Parametric Memory during training but disables it at inference; \textbf{Train \& Inference} reactivates the inner loop on the prompt.}
    \label{tab:longbench}
    \resizebox{\textwidth}{!}{
    \begin{tabular}{l cccccccccccc}
        \toprule
        \textbf{Method} & \textbf{TriviaQA} & \textbf{PR-en} & \textbf{PR-zh} & \textbf{MF-en} & \textbf{MF-zh} & \textbf{2WikiQA} & \textbf{DuReader} & \textbf{MultiNews} & \textbf{TREC} & \textbf{SAMSum} & \textbf{LCC} & \textbf{RB-P} \\
        \midrule
        Standard MDLM          & $55.29$ & $50.32$ & $74.50$ & $29.22$ & $42.47$ & $21.85$ & $22.13$ & $23.52$ & $70.00$ & $13.59$ & $62.55$ & $54.65$ \\
        MemDLM (Train-Only)        & $87.74$ & $54.36$ & $86.29$ & $31.44$ & $42.87$ & $22.43$ & $22.19$ & $26.35$ & $70.50$ & $15.80$ & $64.20$ & $55.05$ \\
        MemDLM (Train \& Inference)& $\mathbf{87.77}$ & $\mathbf{54.69}$ & $\mathbf{87.38}$ & $\mathbf{31.97}$ & $\mathbf{43.28}$ & $\mathbf{22.61}$ & $\mathbf{22.34}$ & $\mathbf{26.70}$ & $\mathbf{71.00}$ & $\mathbf{16.14}$ & $\mathbf{64.25}$ & $\mathbf{55.23}$ \\
        \bottomrule
    \end{tabular}
    }
\end{table}

\subsection{Understanding MemDLM During Training}
\label{sec:exp_training}

\begin{wrapfigure}{r}{0.42\linewidth}
    \centering
    \vspace{-0.5cm}
    \includegraphics[width=\linewidth]{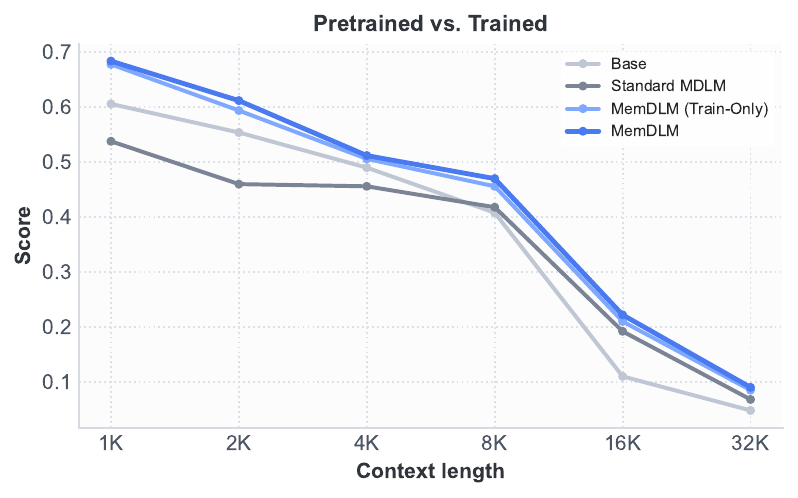}
    \caption{Comparison with the untuned pretrained \texttt{LLaDA-MoE-7B-A1B-Base} model across context lengths. 
    }
    \label{fig:base_compare_analysis}
    \vspace{-0.5cm}
\end{wrapfigure}

\begin{figure*}[t]
    \centering
    \includegraphics[width=\textwidth]{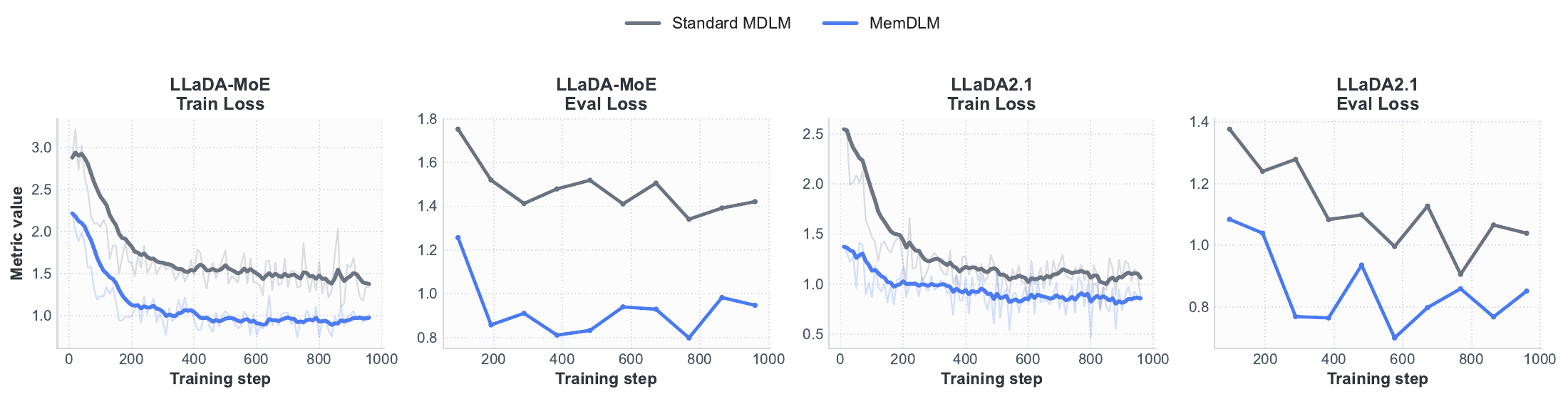}
    \caption{Training dynamics on LLaDA-MoE and LLaDA2.1. Faint train-loss curves show raw values and bold curves show smoothed trends. Across both backbones, MemDLM converges faster and reaches lower train and evaluation loss.}
    \label{fig:training_curves}
    \vspace{-0.5cm}
\end{figure*}

\Cref{fig:training_curves} shows that MemDLM descends more rapidly in training loss and maintains lower evaluation loss throughout training on both backbones. This supports the view that Bi-level Optimization with fast weights improves the learned base model itself rather than acting only as an inference-time mechanism.
We also compare against the untuned \texttt{LLaDA-MoE-7B-A1B-Base} model to understand how training changes pretrained long-context behavior. \Cref{fig:base_compare_analysis} shows that Standard MDLM fine-tuning does not uniformly preserve this capability: it drops below the base model at 1K and 2K, even though it improves at longer contexts. MemDLM instead improves consistently over both the pretrained base and the Standard MDLM-trained model across 1K--32K.

\begin{figure*}[t]
    \centering
    \vspace{-0.8cm}
    \begin{minipage}[t]{0.49\textwidth}
        \centering
        \includegraphics[width=\linewidth]{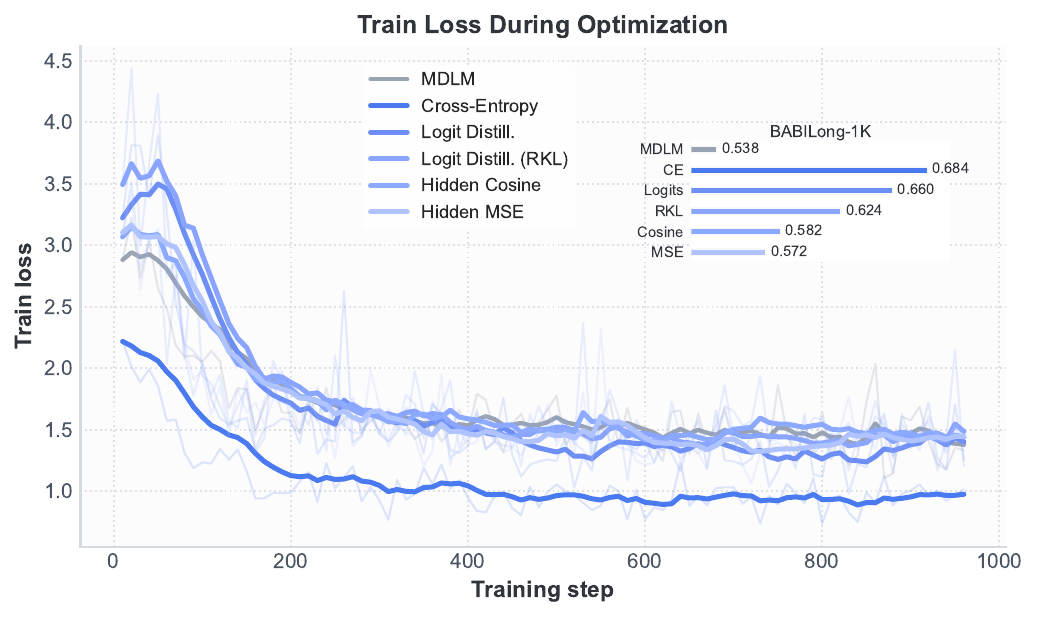}
        \captionof{figure}{Inner-loop supervision analysis on the LLaDA-MoE, evaluated on BABILong-1K. 
        }
        \label{fig:inner_loss_analysis}
    \end{minipage}\hfill
    \begin{minipage}[t]{0.49\textwidth}
        \centering
        \includegraphics[width=\linewidth]{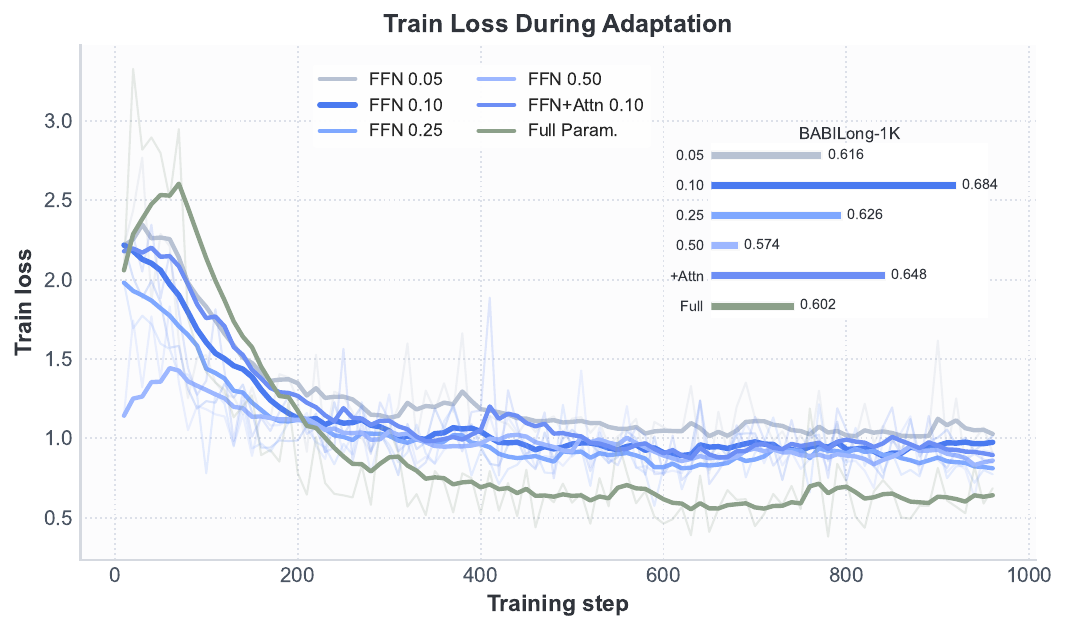}
        \captionof{figure}{Adaptation scope analysis on the LLaDA-MoE, evaluated on BABILong-1K. 
        }
        \label{fig:adapt_scope_analysis}
    \end{minipage}
    \vspace{-0.5cm}
\end{figure*}

\paragraph{Inner-loop supervision.}
We next study which supervision best writes useful trajectory information into the fast weights. Beyond the default cross-entropy objective, we test logit distillation with Kullback-Leibler (KL)~\cite{hinton2015distilling} or reverse-KL divergence and hidden-state distillation with cosine or MSE losses. These variants are a form of \emph{self-distillation}: teacher and student are different views of the same model under different information states. Both branches use the same underlying model with the current fast-weight state, but the teacher is evaluated under \texttt{no\_grad}; in the progressive setting it also sees a later denoising state and therefore more revealed context. \Cref{fig:inner_loss_analysis} shows that MemDLM remains trainable under all of these choices, indicating that the memory-writing mechanism is not tied to a single loss. Cross-entropy still performs best on BABILong-1K ($0.684$), ahead of KL ($0.660$), reverse-KL ($0.624$), cosine ($0.582$), and MSE ($0.572$).

\paragraph{Adaptation scope.}
We also study where the inner-loop updates should be applied. \Cref{fig:adapt_scope_analysis} shows that stronger inner-loop optimization does not necessarily imply better downstream adaptation: full-parameter updates achieve the lowest train loss, yet underperform a much more restricted FFN-only update. Restricting the inner loop to FFN modules in the last $10\%$ of layers yields the best BABILong-1K score ($0.684$), outperforming shallower adaptation ($0.616$ at $5\%$), broader adaptation ($0.626$ at $25\%$, $0.574$ at $50\%$), FFN+attention updates at the same $10\%$ scope ($0.648$), and full-parameter adaptation ($0.602$). Effective Parametric Memory therefore depends not only on capacity, but also on where the update is written.

Additional analyses of gradient normalization and pre-anchor design are deferred to \cref{app:extra_training_ablation}.

\subsection{Understanding MemDLM During Inference}
\label{sec:exp_inference}


\begin{wrapfigure}{r}{0.42\linewidth}
\centering
\vspace{-0.5cm}
\includegraphics[width=\linewidth]{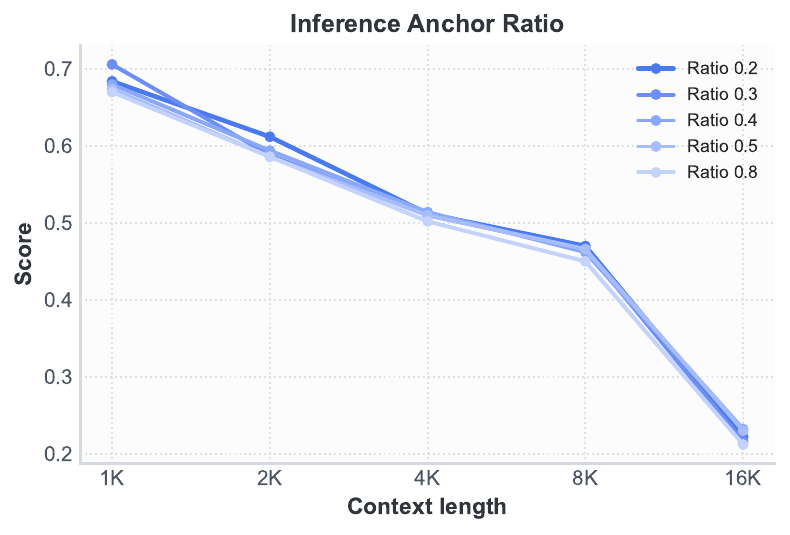}
\caption{\textbf{Sensitivity to the inference anchor ratio.} 
}
\label{fig:inference_anchor_analysis}
\vspace{-0.7cm}
\end{wrapfigure}

Although the main effect of MemDLM appears during training, inference still introduces meaningful design choices. Here we study how the inner loop should be used at inference time and how sensitive it is to the synthetic anchor construction. Our current procedure applies the inner loop to the prompt before generation; adapting during decoding is left to future work.

At inference time, the anchor state is a design choice, which we parameterize by the target mask ratio of the adapted prompt state. \Cref{fig:inference_anchor_analysis} shows that the method is relatively insensitive to this hyperparameter: ratios from $0.2$ to $0.8$ exhibit the same qualitative trend as context length increases, and their scores remain close throughout the full 1K--16K range. Even at 16K, results stay tightly grouped between $0.212$ and $0.232$. We therefore use $0.2$ as the default not because it is uniquely optimal, but because it is a simple operating point in a fairly flat design space.

One possible reason for this low sensitivity is the bidirectional nature of DLM denoising. When the inner loss is computed, the model can attend to all tokens in the corrupted prompt, so changing whether a token is treated as observed input or as a supervised target does not fully remove its information from the local computation. 

\subsection{Ablation of Core Design Choices}
\label{sec:exp_ablation}

\begin{wrapfigure}{r}{0.42\linewidth}
\centering
\vspace{-1cm}
\includegraphics[width=\linewidth]{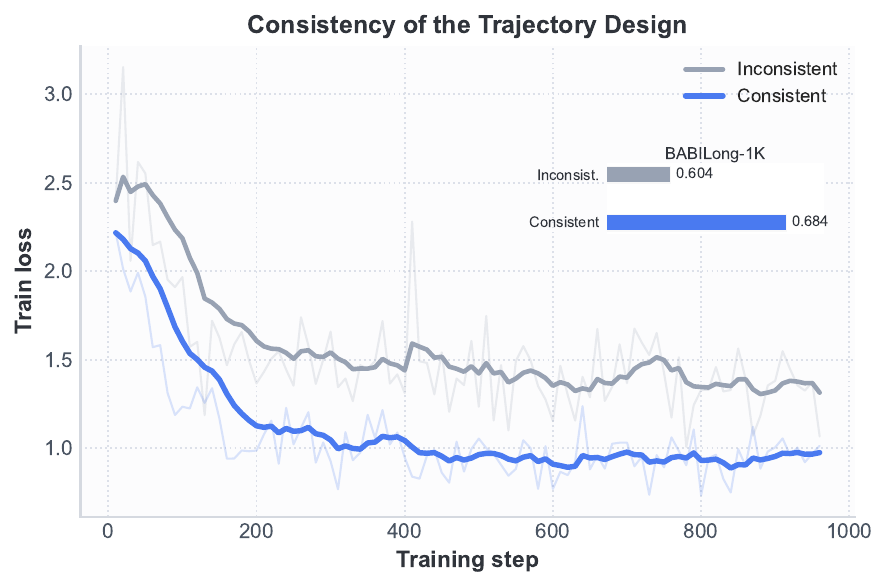}
\caption{\textbf{Consistency of the trajectory.}
}
\label{fig:consistency_analysis}
\vspace{-0.7cm}
\end{wrapfigure}

We also perform ablations that test which components of MemDLM are necessary for the method to work.

\textbf{Consistency of the trajectory design.}
One central hypothesis of MemDLM is that the inner loop should remain consistent with the anchor-centered outer objective. \Cref{fig:consistency_analysis} compares our default design against an inconsistent progressive-memory variant and shows a clear optimization gap: the consistent trajectory converges to substantially lower training loss, and this carries over to downstream retrieval, improving BABILong-1K from $0.604$ to $0.684$. 

\begin{figure*}[t]
\centering
\vspace{-0.1cm}
\begin{minipage}[t]{0.49\textwidth}
\centering
\includegraphics[width=\linewidth]{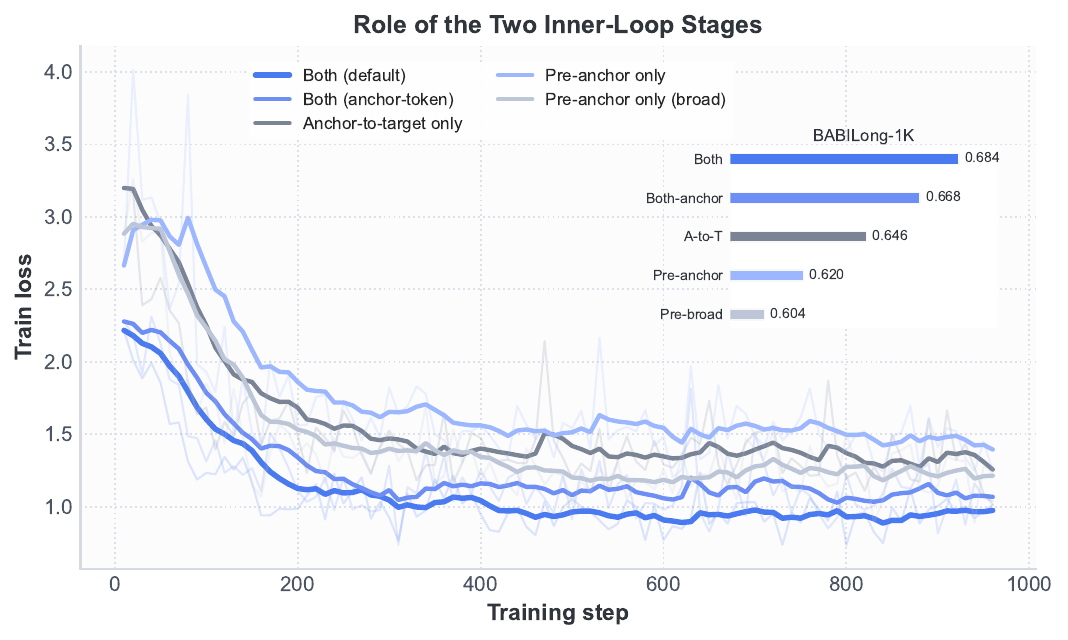}
\captionof{figure}{\textbf{Role of the two inner-loop stages.} 
}
\label{fig:two_stage_analysis}
\end{minipage}\hfill
\begin{minipage}[t]{0.49\textwidth}
\centering
\includegraphics[width=\linewidth]{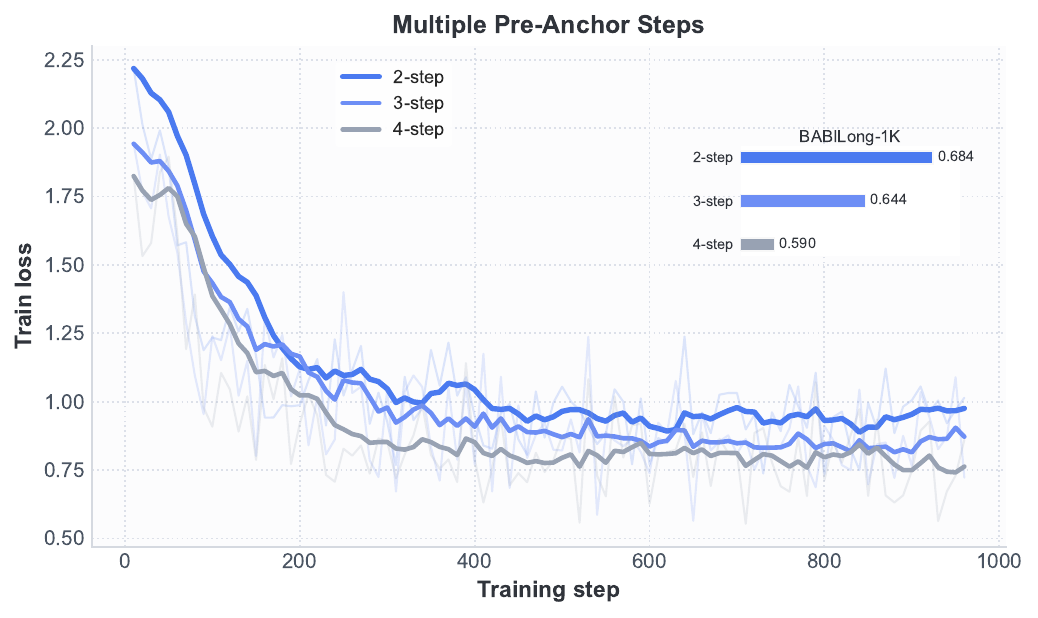}
\captionof{figure}{\textbf{Multiple pre-anchor steps.} 
}
\label{fig:multi_steps_analysis}
\end{minipage}
\vspace{-0.5cm}
\end{figure*}

\textbf{Role of the two inner-loop stages.}
We ablate the two-stage inner loop by using only the pre-anchor stage or only the anchor-to-target stage. \Cref{fig:two_stage_analysis} shows that neither stage alone is sufficient: using only the anchor-to-target stage reaches $0.646$, while using only the pre-anchor stage with anchor-token-only supervision reaches $0.620$. Combining both stages is better, but the exact pre-stage target also matters. With both stages active, restricting the pre-anchor loss to anchor-token-only supervision reaches $0.668$, while our default broader clean-target supervision performs best at $0.684$. Thus the default pre-anchor objective is slightly richer than the idealized stagewise factorization in \cref{sec:method}, but provides a better first-stage update for the subsequent anchor-to-target refinement.

\textbf{Multiple pre-anchor steps.}
Finally, we explore whether multiple pre-anchor steps further improve performance. \Cref{fig:multi_steps_analysis} shows a clear divergence between inner-loop optimization and downstream utility: increasing the number of pre-anchor steps from the default 2-step design to 3-step and 4-step variants steadily lowers the training loss, but the final BABILong-1K score drops from $0.684$ to $0.644$ and then to $0.590$. The current two-stage design therefore appears sufficient for capturing the local trajectory information that matters, and lower inner-loop loss alone is not a reliable proxy.

\section{Related Work}

MemDLM lies at the intersection of diffusion language modeling, fast-weight memory, bi-level adaptation, and inference-time adaptation.

\paragraph{Diffusion language models and the training-inference gap.}
Recent diffusion language models show that masked denoising can support high-quality text generation and flexible infilling, making DLMs a compelling alternative to standard auto-regressive decoding~\cite{austin2021structured, sahoo2024simple, lou2023discrete, shi2024simplified, ou2024your, ye2025dream,zheng2024masked, campbell2022continuous, sun2022score, meng2022concrete, zhen2026dllm, wang2026top}. Several recent works explicitly target the training-inference discrepancy in diffusion decoding. MDPO uses progressive, inference-aligned remasking schedules~\cite{he2025mdpo}; trajectory-aware reinforcement learning (RL) methods optimize the denoising path as a sequential decision process~\cite{wang2025revolutionizing, huang2025reinforcing}; and planner-alignment methods reweight training using the model's own confidence or self-planning signal~\cite{peng2025planner}. MemDLM is motivated by the same mismatch, but addresses it through explicit inner-loop simulation that writes local denoising trajectory information into fast weights during training.

\paragraph{Long-context attention and its limitations.}
As context length grows, dense softmax attention must spread probability mass across all tokens, making it harder to focus on task-relevant information. This has been studied from several angles: Scalable-Softmax replaces the standard softmax with a length-aware normalization~\cite{nakanishi2025scalable}; DySCO dynamically rescales attention during decoding~\cite{ye2026dysco}; attention sinks show that early tokens accumulate disproportionate mass regardless of relevance~\cite{xiao2023efficient}; and positional analyses reveal systematic retrieval failures in the middle of long contexts~\cite{liu2024lost}. These approaches modify the attention mechanism or decoding procedure itself. MemDLM is complementary: rather than changing how attention operates, it reduces the memorization pressure on attention by introducing a parameter-space memory channel during training.

\paragraph{Fast weights and parametric memory.}
The idea that neural networks can store short-lived, sample-specific information in parameters rather than only in activations has a long history in the fast-weights literature~\cite{tieleman2009using, ba2016using, hinton1987using, zhao2026fast}. Related memory-based adaptation methods, many developed in auto-regressive or modern LLM settings, further show that local weight updates can act as parametric memory and enable rapid adaptation from context~\cite{sprechmann2018memory, tack2024online, meng2022mass, mitchell2021fast, memoryllm, wang2024self, padmanabhan2023propagating}. MemDLM is closely connected to this perspective: its fast weights act as a transient parametric memory of a local denoising trajectory, formed directly through inner-loop gradient updates rather than through an external memory module or cache.

\paragraph{Meta-learning and Bi-level Optimization.}
MemDLM relates to meta-learning methods that combine inner-loop adaptation with an outer-loop objective~\cite{thrun1998learning, finn2017model, nichol2018first, vinyals2016matching, snell2017prototypical, santoro2016meta, rajeswaran2019meta, garg2022can}. As in these approaches, our method optimizes base parameters so that a small number of fast updates becomes useful at deployment time. The difference is that our inner loop does not adapt across task episodes in the few-shot sense; instead, it internalizes the local denoising trajectory of each sample.

\paragraph{Test-time training.}
Finally, MemDLM is related to test-time training methods that update model behavior on the fly using unlabeled or self-supervised signals~\cite{pmlr-v119-sun20b, xiong2026scaling, pei2025scope, wang2021tent, zhang2025test, zuo2025ttrl, tandon2025end, zweiger2025self}. TTT-E2E frames long-context modeling as continual test-time learning, using the same next-token objective at training and deployment time so that incoming context can be compressed into the weights during inference~\cite{tandon2025end}. SEAL studies self-adapting language models that generate their own update directives or synthetic supervision and then perform persistent weight updates under a reward-driven loop~\cite{zweiger2025self}. 
This connection is most visible when we re-enable the inner loop at inference time, but our results show that the main gains already emerge from memory-aware training with inference-time adaptation providing only an additional prompt-specific refinement.
\section{Limitations}

MemDLM introduces extra training overhead from the inner loop (${\sim}2\times$ wall-clock under the default configuration; see \cref{app:training_workload}), although we mitigate this with parameter-efficient adapters and restricted update scope. Our experiments are also limited to two DLM backbones and one instruction-tuning dataset, so generalization to other architectures, pretraining-stage optimization, and longer training contexts remains to be tested.
Inference-time adaptation also adds prompt-side latency, which may be undesirable in latency-critical settings.
Extending the inner loop beyond prompt adaptation and into decoding is left for future work.

\section{Conclusion}

We introduced \textbf{MemDLM}, a memory-aware training framework for diffusion language models based on Bi-level Optimization and fast weights as Parametric Memory. Our main finding is that simulating denoising trajectories during training changes what the base model learns: by offloading part of the memorization burden from token-space attention to parameter space, MemDLM improves optimization and strengthens long-context performance even in the \textit{Train-Only} setting. Re-enabling the inner loop at inference provides an additional prompt-specific adaptation effect that we interpret as \emph{in-weight retrieval}. Overall, introducing a parameter-space memory channel appears to be a promising direction for more robust long-context DLMs.


\bibliographystyle{unsrt}
\bibliography{ref/main}

@article{ye2026dysco,
  title={DySCO: Dynamic Attention-Scaling Decoding for Long-Context LMs},
  author={Ye, Xi and Zhang, Wuwei and Yin, Fangcong and Yen, Howard and Chen, Danqi},
  journal={arXiv preprint arXiv:2602.22175},
  year={2026}
}

@article{nakanishi2025scalable,
  title={Scalable-softmax is superior for attention},
  author={Nakanishi, Ken M},
  journal={arXiv preprint arXiv:2501.19399},
  year={2025}
}

@article{xiao2023efficient,
  title={Efficient streaming language models with attention sinks},
  author={Xiao, Guangxuan and Tian, Yuandong and Chen, Beidi and Han, Song and Lewis, Mike},
  journal={arXiv preprint arXiv:2309.17453},
  year={2023}
}

@article{liu2024lost,
  title={Lost in the middle: How language models use long contexts},
  author={Liu, Nelson F and Lin, Kevin and Hewitt, John and Paranjape, Ashwin and Bevilacqua, Michele and Petroni, Fabio and Liang, Percy},
  journal={Transactions of the association for computational linguistics},
  volume={12},
  pages={157--173},
  year={2024}
}

@article{hinton2015distilling,
  title={Distilling the knowledge in a neural network},
  author={Hinton, Geoffrey and Vinyals, Oriol and Dean, Jeff},
  journal={arXiv preprint arXiv:1503.02531},
  year={2015}
}

@article{ye2025dream,
  title={Dream 7B: Diffusion Large Language Models},
  author={Ye, Jiacheng and Xie, Zhihui and Zheng, Lin and Gao, Jiahui and Wu, Zirui and Jiang, Xin and Li, Zhenguo and Kong, Lingpeng},
  journal={arXiv preprint arXiv:2508.15487},
  year={2025}
}

@article{pei2025scope,
  title={Scope: Prompt evolution for enhancing agent effectiveness},
  author={Pei, Zehua and Zhen, Hui-Ling and Kai, Shixiong and Pan, Sinno Jialin and Wang, Yunhe and Yuan, Mingxuan and Yu, Bei},
  journal={arXiv preprint arXiv:2512.15374},
  year={2025}
}

@article{zhen2026dllm,
  title={DLLM Agent: See Farther, Run Faster},
  author={Zhen, Huiling and Lin, Weizhe and Liu, Renxi and Han, Kai and Li, Yiming and Tian, Yuchuan and Chen, Hanting and Li, Xiaoguang and Li, Xiaosong and Chen, Chen and others},
  journal={arXiv preprint arXiv:2602.07451},
  year={2026}
}

@article{xiong2026scaling,
  title={Scaling Search-Augmented LLM Reasoning via Adaptive Information Control},
  author={Xiong, Siheng and Gungordu, Oguzhan and Johnson, Blair and Kerce, James C and Fekri, Faramarz},
  journal={arXiv preprint arXiv:2602.01672},
  year={2026}
}

@inproceedings{bai2024longbench,
  title={Longbench: A bilingual, multitask benchmark for long context understanding},
  author={Bai, Yushi and Lv, Xin and Zhang, Jiajie and Lyu, Hongchang and Tang, Jiankai and Huang, Zhidian and Du, Zhengxiao and Liu, Xiao and Zeng, Aohan and Hou, Lei and others},
  booktitle={Proceedings of the 62nd annual meeting of the association for computational linguistics (volume 1: Long papers)},
  pages={3119--3137},
  year={2024}
}

@article{kuratov2024babilong,
  title={Babilong: Testing the limits of llms with long context reasoning-in-a-haystack},
  author={Kuratov, Yuri and Bulatov, Aydar and Anokhin, Petr and Rodkin, Ivan and Sorokin, Dmitry and Sorokin, Artyom and Burtsev, Mikhail},
  journal={Advances in Neural Information Processing Systems},
  volume={37},
  pages={106519--106554},
  year={2024}
}

@article{hsieh2024ruler,
  title={RULER: What's the real context size of your long-context language models?},
  author={Hsieh, Cheng-Ping and Sun, Simeng and Kriman, Samuel and Acharya, Shantanu and Rekesh, Dima and Jia, Fei and Zhang, Yang and Ginsburg, Boris},
  journal={arXiv preprint arXiv:2404.06654},
  year={2024}
}

@article{loshchilov2017decoupled,
  title={Decoupled weight decay regularization},
  author={Loshchilov, Ilya and Hutter, Frank},
  journal={arXiv preprint arXiv:1711.05101},
  year={2017}
}

@misc{long-alpaca,
  author = {Yukang Chen and Shaozuo Yu and Shengju Qian and Haotian Tang and Xin Lai and Zhijian Liu and Song Han and Jiaya Jia},
  title = {Long Alpaca: Long-context Instruction-following models},
  year = {2023},
  publisher = {GitHub},
  journal = {GitHub repository},
  howpublished = {\url{https://github.com/dvlab-research/LongLoRA}},
}

@article{bie2026llada2,
  title={LLaDA2. 1: Speeding Up Text Diffusion via Token Editing},
  author={Bie, Tiwei and Cao, Maosong and Cao, Xiang and Chen, Bingsen and Chen, Fuyuan and Chen, Kun and Du, Lun and Feng, Daozhuo and Feng, Haibo and Gong, Mingliang and others},
  journal={arXiv preprint arXiv:2602.08676},
  year={2026}
}

@article{zhu2025llada,
  title={Llada-moe: A sparse moe diffusion language model},
  author={Zhu, Fengqi and You, Zebin and Xing, Yipeng and Huang, Zenan and Liu, Lin and Zhuang, Yihong and Lu, Guoshan and Wang, Kangyu and Wang, Xudong and Wei, Lanning and others},
  journal={arXiv preprint arXiv:2509.24389},
  year={2025}
}

@misc{zhou2026dllm,
      title={dLLM: Simple Diffusion Language Modeling}, 
      author={Zhanhui Zhou and Lingjie Chen and Hanghang Tong and Dawn Song},
      year={2026},
      eprint={2602.22661},
      archivePrefix={arXiv},
      primaryClass={cs.CL},
      url={https://arxiv.org/abs/2602.22661}, 
}

@misc{eval-harness,
  author       = {Gao, Leo and Tow, Jonathan and Abbasi, Baber and Biderman, Stella and Black, Sid and DiPofi, Anthony and Foster, Charles and Golding, Laurence and Hsu, Jeffrey and Le Noac'h, Alain and Li, Haonan and McDonell, Kyle and Muennighoff, Niklas and Ociepa, Chris and Phang, Jason and Reynolds, Laria and Schoelkopf, Hailey and Skowron, Aviya and Sutawika, Lintang and Tang, Eric and Thite, Anish and Wang, Ben and Wang, Kevin and Zou, Andy},
  title        = {The Language Model Evaluation Harness},
  month        = 07,
  year         = 2024,
  publisher    = {Zenodo},
  version      = {v0.4.3},
  doi          = {10.5281/zenodo.12608602},
  url          = {https://zenodo.org/records/12608602}
}

@article{zhang2025test,
  title={Test-time training done right},
  author={Zhang, Tianyuan and Bi, Sai and Hong, Yicong and Zhang, Kai and Luan, Fujun and Yang, Songlin and Sunkavalli, Kalyan and Freeman, William T and Tan, Hao},
  journal={arXiv preprint arXiv:2505.23884},
  year={2025}
}

@article{zuo2025ttrl,
  title={Ttrl: Test-time reinforcement learning},
  author={Zuo, Yuxin and Zhang, Kaiyan and Sheng, Li and Qu, Shang and Cui, Ganqu and Zhu, Xuekai and Li, Haozhan and Zhang, Yuchen and Long, Xinwei and Hua, Ermo and others},
  journal={arXiv preprint arXiv:2504.16084},
  year={2025}
}

@article{zweiger2025self,
  title={Self-adapting language models},
  author={Zweiger, Adam and Pari, Jyothish and Guo, Han and Aky{\"u}rek, Ekin and Kim, Yoon and Agrawal, Pulkit},
  journal={arXiv preprint arXiv:2506.10943},
  year={2025}
}

@article{tandon2025end,
  title={End-to-end test-time training for long context},
  author={Tandon, Arnuv and Dalal, Karan and Li, Xinhao and Koceja, Daniel and R{\o}d, Marcel and Buchanan, Sam and Wang, Xiaolong and Leskovec, Jure and Koyejo, Sanmi and Hashimoto, Tatsunori and others},
  journal={arXiv preprint arXiv:2512.23675},
  year={2025}
}

@inproceedings{wang2021tent,
  title={Tent: Fully Test-Time Adaptation by Entropy Minimization},
  author={Wang, Dequan and Shelhamer, Evan and Liu, Shaoteng and Olshausen, Bruno and Darrell, Trevor},
  booktitle={International Conference on Learning Representations},
  year={2021},
  url={https://openreview.net/forum?id=uXl3bZLkr3c}
}

@InProceedings{pmlr-v119-sun20b,
  title = 	 {Test-Time Training with Self-Supervision for Generalization under Distribution Shifts},
  author =       {Sun, Yu and Wang, Xiaolong and Liu, Zhuang and Miller, John and Efros, Alexei and Hardt, Moritz},
  booktitle = 	 {Proceedings of the 37th International Conference on Machine Learning},
  pages = 	 {9229--9248},
  year = 	 {2020},
  editor = 	 {III, Hal Daumé and Singh, Aarti},
  volume = 	 {119},
  series = 	 {Proceedings of Machine Learning Research},
  month = 	 {13--18 Jul},
  publisher =    {PMLR},
  url = 	 {https://proceedings.mlr.press/v119/sun20b.html}
}

@article{garg2022can,
  title={What can transformers learn in-context? a case study of simple function classes},
  author={Garg, Shivam and Tsipras, Dimitris and Liang, Percy S and Valiant, Gregory},
  journal={Advances in neural information processing systems},
  volume={35},
  pages={30583--30598},
  year={2022}
}

@article{rajeswaran2019meta,
  title={Meta-learning with implicit gradients},
  author={Rajeswaran, Aravind and Finn, Chelsea and Kakade, Sham M and Levine, Sergey},
  journal={Advances in neural information processing systems},
  volume={32},
  year={2019}
}

@inproceedings{santoro2016meta,
  title={Meta-learning with memory-augmented neural networks},
  author={Santoro, Adam and Bartunov, Sergey and Botvinick, Matthew and Wierstra, Daan and Lillicrap, Timothy},
  booktitle={International conference on machine learning},
  pages={1842--1850},
  year={2016},
  organization={PMLR}
}

@article{snell2017prototypical,
  title={Prototypical networks for few-shot learning},
  author={Snell, Jake and Swersky, Kevin and Zemel, Richard},
  journal={Advances in neural information processing systems},
  volume={30},
  year={2017}
}

@article{vinyals2016matching,
  title={Matching networks for one shot learning},
  author={Vinyals, Oriol and Blundell, Charles and Lillicrap, Timothy and Wierstra, Daan and others},
  journal={Advances in neural information processing systems},
  volume={29},
  year={2016}
}

@incollection{thrun1998learning,
  title={Learning to learn: Introduction and overview},
  author={Thrun, Sebastian and Pratt, Lorien},
  booktitle={Learning to learn},
  pages={3--17},
  year={1998},
  publisher={Springer}
}

@article{nichol2018first,
  title={On first-order meta-learning algorithms},
  author={Nichol, Alex and Achiam, Joshua and Schulman, John},
  journal={arXiv preprint arXiv:1803.02999},
  year={2018}
}

@inproceedings{finn2017model,
  title={Model-agnostic meta-learning for fast adaptation of deep networks},
  author={Finn, Chelsea and Abbeel, Pieter and Levine, Sergey},
  booktitle={International conference on machine learning},
  pages={1126--1135},
  year={2017},
  organization={PMLR}
}

@article{zhao2026fast,
  title={Fast-weight Product Key Memory},
  author={Zhao, Tianyu and Jones, Llion},
  journal={arXiv preprint arXiv:2601.00671},
  year={2026}
}

@inproceedings{memoryllm,
  author       = {Yu Wang and
                  Yifan Gao and
                  Xiusi Chen and
                  Haoming Jiang and
                  Shiyang Li and
                  Jingfeng Yang and
                  Qingyu Yin and
                  Zheng Li and
                  Xian Li and
                  Bing Yin and
                  Jingbo Shang and
                  Julian J. McAuley},
  title        = {{MEMORYLLM:} Towards Self-Updatable Large Language Models},
  booktitle    = {Forty-first International Conference on Machine Learning, {ICML} 2024,
                  Vienna, Austria, July 21-27, 2024},
  publisher    = {OpenReview.net},
  year         = {2024},
  url          = {https://openreview.net/forum?id=p0lKWzdikQ},
  bibsource    = {dblp computer science bibliography, https://dblp.org}
}

@article{padmanabhan2023propagating,
  title={Propagating knowledge updates to lms through distillation},
  author={Padmanabhan, Shankar and Onoe, Yasumasa and Zhang, Michael and Durrett, Greg and Choi, Eunsol},
  journal={Advances in Neural Information Processing Systems},
  volume={36},
  pages={47124--47142},
  year={2023}
}

@article{wang2024self,
  title={Self-updatable large language models by integrating context into model parameters},
  author={Wang, Yu and Liu, Xinshuang and Chen, Xiusi and O'Brien, Sean and Wu, Junda and McAuley, Julian},
  journal={arXiv preprint arXiv:2410.00487},
  year={2024}
}

@article{mitchell2021fast,
  title={Fast model editing at scale},
  author={Mitchell, Eric and Lin, Charles and Bosselut, Antoine and Finn, Chelsea and Manning, Christopher D},
  journal={arXiv preprint arXiv:2110.11309},
  year={2021}
}

@article{meng2022mass,
  title={Mass-editing memory in a transformer},
  author={Meng, Kevin and Sharma, Arnab Sen and Andonian, Alex and Belinkov, Yonatan and Bau, David},
  journal={arXiv preprint arXiv:2210.07229},
  year={2022}
}

@article{tack2024online,
  title={Online adaptation of language models with a memory of amortized contexts},
  author={Tack, Jihoon and Kim, Jaehyung and Mitchell, Eric and Shin, Jinwoo and Teh, Yee Whye and Schwarz, Jonathan Richard},
  journal={Advances in Neural Information Processing Systems},
  volume={37},
  pages={130109--130135},
  year={2024}
}

@article{sprechmann2018memory,
  title={Memory-based parameter adaptation},
  author={Sprechmann, Pablo and Jayakumar, Siddhant M and Rae, Jack W and Pritzel, Alexander and Badia, Adria Puigdomenech and Uria, Benigno and Vinyals, Oriol and Hassabis, Demis and Pascanu, Razvan and Blundell, Charles},
  journal={arXiv preprint arXiv:1802.10542},
  year={2018}
}

@inproceedings{hinton1987using,
  title={Using fast weights to deblur old memories},
  author={Hinton, Geoffrey E and Plaut, David C},
  booktitle={Proceedings of the ninth annual conference of the Cognitive Science Society},
  pages={177--186},
  year={1987}
}

@article{ba2016using,
  title={Using fast weights to attend to the recent past},
  author={Ba, Jimmy and Hinton, Geoffrey E and Mnih, Volodymyr and Leibo, Joel Z and Ionescu, Catalin},
  journal={Advances in neural information processing systems},
  volume={29},
  year={2016}
}

@inproceedings{tieleman2009using,
  title={Using fast weights to improve persistent contrastive divergence},
  author={Tieleman, Tijmen and Hinton, Geoffrey},
  booktitle={Proceedings of the 26th annual international conference on machine learning},
  pages={1033--1040},
  year={2009}
}

@article{peng2025planner,
  title={Planner Aware Path Learning in Diffusion Language Models Training},
  author={Peng, Fred Zhangzhi and Bezemek, Zachary and Rector-Brooks, Jarrid and Zhang, Shuibai and Zhang, Anru R and Bronstein, Michael and Bose, Avishek Joey and Tong, Alexander},
  journal={arXiv preprint arXiv:2509.23405},
  year={2025}
}

@article{huang2025reinforcing,
  title={Reinforcing the diffusion chain of lateral thought with diffusion language models},
  author={Huang, Zemin and Chen, Zhiyang and Wang, Zijun and Li, Tiancheng and Qi, Guo-Jun},
  journal={arXiv preprint arXiv:2505.10446},
  year={2025}
}

@article{wang2025revolutionizing,
  title={Revolutionizing reinforcement learning framework for diffusion large language models},
  author={Wang, Yinjie and Yang, Ling and Li, Bowen and Tian, Ye and Shen, Ke and Wang, Mengdi},
  journal={arXiv preprint arXiv:2509.06949},
  year={2025}
}

@article{meng2022concrete,
  title={Concrete score matching: Generalized score matching for discrete data},
  author={Meng, Chenlin and Choi, Kristy and Song, Jiaming and Ermon, Stefano},
  journal={Advances in Neural Information Processing Systems},
  volume={35},
  pages={34532--34545},
  year={2022}
}

@article{sun2022score,
  title={Score-based continuous-time discrete diffusion models},
  author={Sun, Haoran and Yu, Lijun and Dai, Bo and Schuurmans, Dale and Dai, Hanjun},
  journal={arXiv preprint arXiv:2211.16750},
  year={2022}
}

@article{campbell2022continuous,
  title={A continuous time framework for discrete denoising models},
  author={Campbell, Andrew and Benton, Joe and De Bortoli, Valentin and Rainforth, Thomas and Deligiannidis, George and Doucet, Arnaud},
  journal={Advances in Neural Information Processing Systems},
  volume={35},
  pages={28266--28279},
  year={2022}
}

@article{zheng2024masked,
  title={Masked diffusion models are secretly time-agnostic masked models and exploit inaccurate categorical sampling},
  author={Zheng, Kaiwen and Chen, Yongxin and Mao, Hanzi and Liu, Ming-Yu and Zhu, Jun and Zhang, Qinsheng},
  journal={arXiv preprint arXiv:2409.02908},
  year={2024}
}

@article{ou2024your,
  title={Your absorbing discrete diffusion secretly models the conditional distributions of clean data},
  author={Ou, Jingyang and Nie, Shen and Xue, Kaiwen and Zhu, Fengqi and Sun, Jiacheng and Li, Zhenguo and Li, Chongxuan},
  journal={arXiv preprint arXiv:2406.03736},
  year={2024}
}

@article{shi2024simplified,
  title={Simplified and generalized masked diffusion for discrete data},
  author={Shi, Jiaxin and Han, Kehang and Wang, Zhe and Doucet, Arnaud and Titsias, Michalis},
  journal={Advances in neural information processing systems},
  volume={37},
  pages={103131--103167},
  year={2024}
}

@article{austin2021structured,
  title={Structured denoising diffusion models in discrete state-spaces},
  author={Austin, Jacob and Johnson, Daniel D and Ho, Jonathan and Tarlow, Daniel and Van Den Berg, Rianne},
  journal={Advances in neural information processing systems},
  volume={34},
  pages={17981--17993},
  year={2021}
}

@article{he2025mdpo,
  title={Mdpo: Overcoming the training-inference divide of masked diffusion language models},
  author={He, Haoyu and Renz, Katrin and Cao, Yong and Geiger, Andreas},
  journal={arXiv preprint arXiv:2508.13148},
  year={2025}
}

@article{lou2023discrete,
  title={Discrete diffusion modeling by estimating the ratios of the data distribution},
  author={Lou, Aaron and Meng, Chenlin and Ermon, Stefano},
  journal={arXiv preprint arXiv:2310.16834},
  year={2023}
}

@article{sahoo2024simple,
  title={Simple and effective masked diffusion language models},
  author={Sahoo, Subham S and Arriola, Marianne and Schiff, Yair and Gokaslan, Aaron and Marroquin, Edgar and Chiu, Justin T and Rush, Alexander and Kuleshov, Volodymyr},
  journal={Advances in Neural Information Processing Systems},
  volume={37},
  pages={130136--130184},
  year={2024}
}

@article{hu2021lora,
  title={Lora: Low-rank adaptation of large language models},
  author={Hu, Edward J and Shen, Yelong and Wallis, Phillip and Allen-Zhu, Zeyuan and Li, Yuanzhi and Wang, Shean and Wang, Lu and Chen, Weizhu},
  journal={arXiv preprint arXiv:2106.09685},
  year={2021}
}

@article{wang2026top,
  title={Top 10 Open Challenges Steering the Future of Diffusion Language Model and Its Variants},
  author={Wang, Yunhe and Han, Kai and Zhen, Huiling and Tian, Yuchuan and Chen, Hanting and Huang, Yongbing and Cui, Yufei and Shu, Yingte and Gao, Shan and Elezi, Ismail and others},
  journal={arXiv preprint arXiv:2601.14041},
  year={2026}
}

@article{paszke2019pytorch,
  title={Pytorch: An imperative style, high-performance deep learning library},
  author={Paszke, Adam and Gross, Sam and Massa, Francisco and Lerer, Adam and Bradbury, James and Chanan, Gregory and Killeen, Trevor and Lin, Zeming and Gimelshein, Natalia and Antiga, Luca and others},
  journal={Advances in neural information processing systems},
  volume={32},
  year={2019}
}

\appendix


\newpage
\section{Additional Experimental Details}
\label{app:exp_setup}

\textbf{Implementation and Baselines.} We implement MemDLM in PyTorch~\cite{paszke2019pytorch} on top of the open-source \texttt{dllm}~\cite{zhou2026dllm} training library and use \texttt{lm-evaluation-harness}~\cite{eval-harness} for downstream evaluation. We study two backbones in the main experiments: \texttt{LLaDA-MoE-7B-A1B-Base}~\cite{zhu2025llada} and \texttt{LLaDA2.1-mini} (16B)~\cite{bie2026llada2}, abbreviated throughout the paper as \textbf{LLaDA-MoE} and \textbf{LLaDA2.1}. Unless otherwise noted, targeted training-stage analyses and core ablations are conducted on the LLaDA-MoE backbone, while the main retrieval and optimization comparisons are reported on both backbones. Our baseline is \textbf{Standard MDLM}~\cite{sahoo2024simple}, which optimizes only the standard denoising objective (equivalent to our outer loop) with a time-dependent reweighting schedule.

\textbf{Training Data and Processing.} We conduct instruction tuning on LongAlpaca~\cite{long-alpaca}, filtering the dataset to sequences of at most $4,096$ tokens for computational efficiency. During training, we apply an asymmetric masking strategy: prompt tokens remain strictly unmasked and are excluded from the loss, while the noise and masking processes are applied only to the response tokens.

\textbf{Hyperparameters, optimization, and compute.} For parameter efficiency, we load the base model in 4-bit quantization and apply Low-Rank Adaptation (LoRA)~\cite{hu2021lora} for the outer loop with rank $r=32$ and $\alpha=64$. The outer loop is optimized using AdamW~\cite{loshchilov2017decoupled} with learning rate $2 \times 10^{-5}$ and a cosine schedule with $0.1$ warmup ratio. For the Parametric Memory mechanism, the inner loop uses a separate transient set of LoRA adapters with the same configuration ($r=32, \alpha=64$). To reduce overhead, the inner loop only updates Feed-Forward Network (FFN) modules in the final fraction of transformer layers. The inner-loop adaptation consists of a single epoch of SGD with learning rate $0.1$ and gradient clipping set to $1.0$.

All training and evaluation use $8$ NVIDIA H200 GPUs. We train with batch size $2$ per device for $5$ epochs; under our default MemDLM setting, wall-clock training is approximately $1$ hour for \texttt{LLaDA-MoE-7B-A1B-Base} and approximately $3$ hours for \texttt{LLaDA2.1-mini}.

\textbf{Evaluation benchmarks.} We evaluate on \textbf{RULER}~\cite{hsieh2024ruler}, \textbf{BABILong}~\cite{kuratov2024babilong}, and \textbf{LongBench}~\cite{bai2024longbench}. Expanded descriptions, including the RULER sub-tasks we report, are given in \cref{app:benchmark_details}.

\textbf{Generation and decoding settings.} Decoding hyperparameters are chosen to match each backbone's reference practice while keeping denoising behavior consistent where noted. For \textbf{LLaDA-MoE}, we use \emph{single-block} generation and set the output length according to each benchmark's requirements. For \textbf{LLaDA2.1}, we use \emph{multi-block} generation. For \emph{both} models, during iterative denoising we generate \emph{one token per step}. Other shared evaluation options (e.g., via \texttt{lm-evaluation-harness}) are held fixed across compared methods for a given backbone.

\section{Details of evaluation benchmarks}
\label{app:benchmark_details}

\textbf{RULER}~\cite{hsieh2024ruler} is a \emph{synthetic} benchmark for long-context models with configurable sequence length and task complexity. In the official benchmark, tasks are grouped into retrieval, multi-hop tracing, aggregation, and question-answering families. We report three representative sub-tasks used in our main table (see also the official RULER repository task table):

\begin{itemize}
    \item \textbf{Multi-Value (MV).} A retrieval-style extension of needle-in-a-haystack in which multiple target values are embedded in long distractor text and the model must return the requested set of values, rather than only one needle.
    \item \textbf{Variable Tracking (VT).} The official RULER task \texttt{variable\_tracking}, where the context contains chains of variable name-binding updates and the model must follow the chain to identify the final binding.
    \item \textbf{Common Words Extraction (CWE).} The official RULER task \texttt{common\_words\_extraction}, where the context is constructed with controlled word frequencies and the model must extract the words that occur commonly under the task specification.
\end{itemize}

\textbf{BABILong}~\cite{kuratov2024babilong} evaluates \emph{reasoning-in-a-haystack}: it extends bAbI-style reasoning to long-context settings by scattering the relevant supporting facts through very long documents. The benchmark contains 20 reasoning tasks, including fact chaining, induction, deduction, counting, and list/set manipulation, and is designed to test whether models can both retrieve and combine distant evidence as context length grows.

\textbf{LongBench}~\cite{bai2024longbench} is a \emph{bilingual}, multi-task benchmark for realistic long-context understanding. Our main table (\cref{tab:longbench}) follows the task names used in the LongBench suite and our evaluation harness: \textbf{TriviaQA}, \textbf{PR-en}, \textbf{PR-zh}, \textbf{MF-en}, \textbf{MF-zh}, \textbf{2WikiQA}, \textbf{DuReader}, \textbf{MultiNews}, \textbf{TREC}, \textbf{SAMSum}, \textbf{LCC}, and \textbf{RB-P} (RepoBench-P). Concretely, this subset covers question answering, passage retrieval, summarization, classification, and code-completion style long-context tasks in both English and Chinese.

\section{Empirical Analysis of Exposure Bias}
\label{app:exposure_bias}

To empirically quantify the train-inference mismatch discussed in \cref{sec:motivation}, we compare the model's loss under two conditions on a validation set of prompt-response pairs. For a given mask ratio corresponding to timestep $t$:

\textbf{Static Condition:} The model predicts masked tokens from a pristine context where the ground-truth response is artificially masked according to the true forward process:
\begin{equation}
    \mathcal{L}_{\text{static}} = \mathbb{E}_{x_0, x_t \sim q(\cdot | x_0)} \left[ -\log p_\theta(x_0 | x_t) \right].
\end{equation}

\textbf{Sequential Condition:} Starting from a $100\%$ masked response, the model iteratively predicts and unmasks tokens using its own predictions until reaching timestep $t$:
\begin{equation}
    \mathcal{L}_{\text{seq}} = \mathbb{E}_{x_0, \hat{x}_t \sim p_\theta} \left[ -\log p_\theta(x_0 | \hat{x}_t) \right].
\end{equation}

We define the \textbf{Exposure Bias Ratio} as $\mathcal{R}_{\text{EB}} = \mathcal{L}_{\text{seq}}/\mathcal{L}_{\text{static}}$. A higher $\mathcal{R}_{\text{EB}}$ indicates more severe exposure bias, meaning the model struggles to denoise its own intermediate representations. 

\begin{figure}[h]
    \centering
    \includegraphics[width=0.5\linewidth]{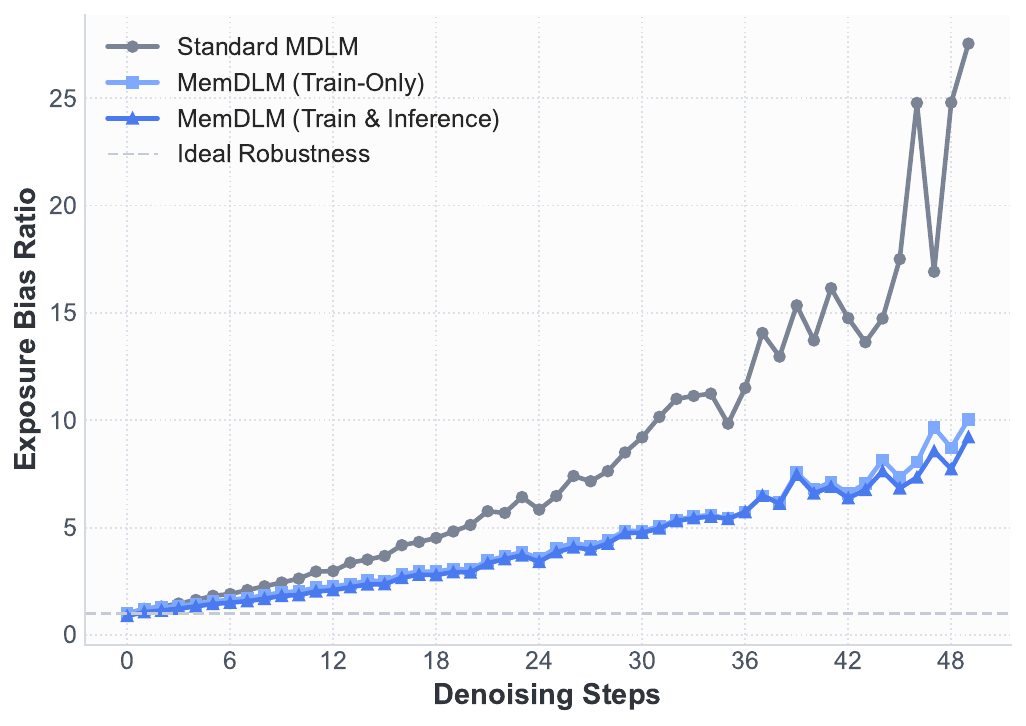}
    \caption{Exposure Bias Ratio ($\mathcal{R}_{\text{EB}}$) across denoising steps. Standard MDLM degrades rapidly, while MemDLM remains substantially flatter.}
    \label{fig:exposure_bias}
\end{figure}

\Cref{fig:exposure_bias} shows that a Standard MDLM exhibits a steeply rising exposure-bias curve. By the end of generation, the sequential loss is substantially higher than the static loss, confirming that standard training leaves the model vulnerable to its own sequential noise. Even in \textit{Train-Only} mode, where the inner loop is disabled at inference, MemDLM exhibits a substantially flatter degradation curve than the baseline, suggesting that the training-time benefit of memory-aware optimization already improves the robustness of the learned base model. Re-enabling the inner loop at inference smooths the curve further, indicating an additional prompt-specific adaptation effect.

\section{Additional Training Ablations}
\label{app:extra_training_ablation}

\paragraph{Gradient normalization in the inner loop.}
Because the inner loop performs rapid task-local adaptation, its behavior can depend on gradient normalization. On LLaDA-MoE / BABILong-1K, local per-parameter normalization with clipping $1.0$ achieves the best score ($0.684$), whereas global normalization degrades performance to $0.632$. Varying the clipping threshold under local normalization has a weaker effect: clipping at $0.5$ or $2.0$ yields $0.630$ and $0.640$, while removing clipping remains competitive at $0.682$. The key choice therefore appears to be \emph{local} normalization rather than the exact clipping threshold.

\paragraph{Pre-anchor design.}
We also investigate the pre-anchor state $x_{t_{\text{pre}}}$. In the anchor-consistent setting, its mask ratio is controlled by a pre-anchor scale hyperparameter $s_{\text{pre}}$, which sets the starting ratio as $\min(1, \max(s_{\text{pre}} \cdot t, t))$ for anchor mask ratio $t$. A scale of $1.5$ performs best ($0.684$), while nearby values $1.75$ and $2.0$ remain competitive ($0.674$ and $0.678$); a smaller scale $1.25$ performs worse ($0.624$). The design is therefore meaningful but not highly fragile once the pre-anchor state is sufficiently noisier than the anchor.

\section{Training Workload Analysis}
\label{app:training_workload}

\Cref{tab:workload} reports wall-clock training time and per-step throughput for Standard MDLM and MemDLM under the default configuration (2-step inner loop, FFN-only LoRA on the last 10\% of layers), as well as selected inner-loop variants. All runs use 8 NVIDIA H200 GPUs, batch size 2 per device, and 960 training steps (5 epochs).

\begin{table}[h]
    \centering
    \caption{Training workload comparison. Wall-clock time and per-step cost are measured from TensorBoard logs. Overhead is relative to the Standard MDLM baseline of each backbone.}
    \label{tab:workload}
    \resizebox{\textwidth}{!}{
    \begin{tabular}{l l r r r r}
        \toprule
        \textbf{Backbone} & \textbf{Configuration} & \textbf{Wall Time (h)} & \textbf{s/step} & \textbf{Overhead} & \textbf{BABILong-1K} \\
        \midrule
        \multirow{2}{*}{LLaDA-MoE} & Standard MDLM & $0.47$ & $1.78$ & $1.0\times$ & $0.538$ \\
         & MemDLM (default) & $0.97$ & $3.67$ & $2.1\times$ & $0.684$ \\
        \midrule
        \multirow{2}{*}{LLaDA2.1} & Standard MDLM & $1.60$ & $6.08$ & $1.0\times$ & $0.646$ \\
         & MemDLM (default) & $3.40$ & $12.87$ & $2.1\times$ & $0.706$ \\
        \midrule
        \multicolumn{6}{l}{\emph{Adaptation scope variants (LLaDA-MoE)}} \\
        \midrule
         & FFN, last 5\% layers & $0.92$ & $3.51$ & $2.0\times$ & $0.616$ \\
         & FFN, last 10\% layers (default) & $0.97$ & $3.67$ & $2.1\times$ & $0.684$ \\
         & FFN+Attention, last 10\% layers & $0.99$ & $3.75$ & $2.1\times$ & $0.648$ \\
         & FFN, last 25\% layers & $1.09$ & $4.14$ & $2.3\times$ & $0.626$ \\
         & FFN, last 50\% layers & $1.31$ & $4.95$ & $2.8\times$ & $0.574$ \\
         & Full parameters & $1.68$ & $6.38$ & $3.6\times$ & $0.602$ \\
        \midrule
        \multicolumn{6}{l}{\emph{Inner-loop stage variants (LLaDA-MoE)}} \\
        \midrule
         & Pre-anchor only (1-step) & $0.81$ & $3.08$ & $1.7\times$ & $0.620$ \\
         & Anchor-to-target only (1-step) & $0.76$ & $2.88$ & $1.6\times$ & $0.646$ \\
         & Both stages, 2-step (default) & $0.97$ & $3.67$ & $2.1\times$ & $0.684$ \\
         & Both stages, 4-step & $1.46$ & $5.55$ & $3.1\times$ & $0.590$ \\
        \bottomrule
    \end{tabular}
    }
\end{table}

Under the default configuration, MemDLM adds approximately $2.1\times$ wall-clock overhead compared to Standard MDLM on both backbones. This overhead is consistent: training LLaDA-MoE takes ${\sim}1$ hour (vs.\ ${\sim}0.5$ hours for the baseline) and LLaDA2.1 takes ${\sim}3.4$ hours (vs.\ ${\sim}1.6$ hours). The per-step cost scales predictably with the number of parameters updated in the inner loop: restricting updates to FFN modules in the last 5--10\% of layers keeps overhead near $2\times$, while full-parameter inner-loop updates raise it to $3.6\times$. Similarly, increasing the number of inner-loop steps from 2 to 4 increases overhead from $2.1\times$ to $3.1\times$. As shown in the main text, the default restricted configuration (FFN, last 10\%, 2-step) achieves the best downstream performance, so the most effective setting is also the most efficient.

\end{document}